\documentclass[letterpaper, 10 pt, conference]{ieeeconf}  
\usepackage{csquotes}
\usepackage{graphicx}
\usepackage[caption=false,font=footnotesize]{subfig} 
\usepackage{xcolor}
\usepackage{pifont}
\usepackage{amsmath, amssymb, mathtools}
\mathtoolsset{showonlyrefs,centercolon} 

\usepackage{balance}
\usepackage[T1]{fontenc}
\usepackage[utf8]{inputenc}
\usepackage{authblk}
\IEEEoverridecommandlockouts       

\title{\LARGE \bf
Zero-Shot Deformation Reconstruction for Soft Robots Using a Flexible Sensor Array and Cage-Based 3D Gaussian Modeling
}

\author[1]{Linrui Shou}
\author[1]{Zilang Chen}
\author[1]{Wenjia Xu}
\author[2]{Yiyue Luo}
\author[1]{Tingyu Cheng}

\affil[1]{University of Notre Dame}
\affil[2]{University of Washington}

\usepackage{cite}

\usepackage[hidelinks]{hyperref}
\begin{document}

\maketitle
\thispagestyle{empty}
\pagestyle{empty}

\begin{figure*}[t]
    \centering
    \includegraphics[width=\linewidth]{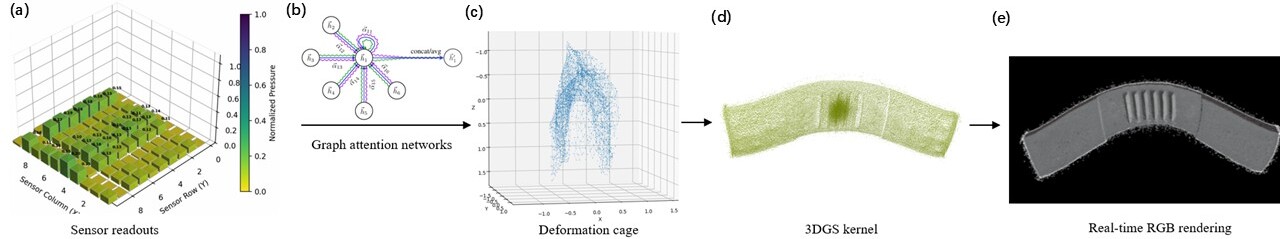}
    \caption{
    Overview of our camera-free framework for real-time reconstruction and RGB rendering of deformable objects from a flexible sensor array.
    (a) Real-time resistance map acquired from the flexible tactile array.
    (b) A Graph Attention Network maps encoded sensor features to cage node displacements.
    (c) Low-dimensional cage representation controlling dense surface deformation.
    (d) Gaussian primitives updated via cage-based interpolation.
    (e) Real-time rendering of the deformed object using Gaussian splatting.
    }
    \label{fig:1}
\end{figure*}
\begin{abstract}
We present a zero-shot deformation reconstruction framework for soft robots that operates without any visual supervision at inference time. In this work, zero-shot deformation reconstruction is defined as the ability to infer object-wide deformations on previously unseen soft robots without collecting object-specific deformation data or performing any retraining during deployment. Our method assumes access to a static geometric proxy of the undeformed object, which can be obtained  from a STL model. During operation, the system relies exclusively on tactile sensing, enabling camera-free deformation inference. The proposed framework integrates a flexible piezoresistive sensor array with a geometry-aware, cage-based 3D Gaussian deformation model. Local tactile measurements are mapped to low-dimensional cage control signals and propagated to dense Gaussian primitives to generate globally consistent shape deformations. A graph attention network regresses cage displacements from tactile input, enforcing spatial smoothness and structural continuity via boundary-aware propagation. Given only a nominal geometric proxy and real-time tactile signals, the system performs zero-shot deformation reconstruction of unseen soft robots in bending and twisting motions, while rendering photorealistic RGB in real time. It achieves 0.67 IoU, 0.65 SSIM, and 3.48 mm Chamfer distance, demonstrating strong zero-shot generalization through explicit coupling of tactile sensing and structured geometric deformation.
\end{abstract}

\section{Introduction}

Accurate three-dimensional (3D) reconstruction of soft robots is a fundamental capability for perception, control, and interaction in complex, constrained, and safety-critical environments.
Applications such as minimally invasive surgery, in situ manipulation, and navigation in occluded or confined spaces critically depend on reliable estimation of deformable robot geometries \cite{huang_endo-4dgs_2024, xin_electrical_2023, chen_robot_2025, luo_tactile_2024}.
Compared with rigid robots, soft robots exhibit continuous deformation, self-occlusion, and virtually infinite degrees of freedom, which render conventional rigid-body modeling and camera-based reconstruction pipelines insufficient or impractical \cite{li_neuralangelo_2023, song_nerfplayer_2023, kerbl_3d_2023}.

To overcome visual limitations, a growing body of work explores camera-free reconstruction of soft robots using embedded sensing.
Electrical impedance tomography (EIT)-based approaches infer volumetric or surface deformation from distributed electrode measurements \cite{xin_electrical_2023, jiang_electrical_2024, chen_multi-touch_2025}.
In parallel, tactile-based and electronic-skin-based systems leverage dense strain or pressure sensing to estimate robot shape or contact states \cite{park_stretchable_2024, mei_high-density_2024, dong_learning-enhanced_2025}.
Recent learning-based methods further demonstrate that such sensing modalities can support reconstruction under a three-dimensional semantic representation, such as dense point clouds or surface geometry, rather than only low-dimensional descriptors like curvature or bending angles \cite{luo_tactile_2024}.

Despite these advances, existing EIT-based and tactile-based reconstruction methods fundamentally rely on robot-specific pre-training and pre-defined sensing configurations \cite{koch_surface_2017, chen_robot_2025, hu_stretchable_2023, park_stretchable_2024, truby_distributed_2020}.
In practice, the sensor layout is manually designed and fixed for a particular soft robot morphology, and the reconstruction model is trained using paired sensor--geometry data collected on that specific robot \cite{koch_surface_2017, huang_3d-vitac_2025, dong_learning-enhanced_2025}.
Consequently, these methods exhibit little to no zero-shot capability: a trained model cannot be directly deployed on an unseen soft robot with different geometry, material distribution, or sensor placement.

The lack of zero-shot generalization stems from three main limitations.
First, there is no efficient mechanism to rapidly obtain an accurate initial 3D model of a soft robot; most sensor-driven pipelines assume a predefined template or rely on external motion-capture and scanning systems during calibration \cite{koch_surface_2017, chen_multi-touch_2025, lun_real-time_2019}, which constrains scalability.
Second, learning-based sensor-to-geometry mappings are tightly coupled to specific robot morphologies and sensor layouts, implicitly encoding robot-dependent assumptions and generalizing poorly to unseen geometries \cite{huang_3d-vitac_2025, park_stretchable_2024, truby_distributed_2020}.
Third, even with relatively dense sensing, reconstruction methods remain largely geometric—using point clouds or surface meshes without explicit enforcement of global smoothness or structural consistency \cite{koch_surface_2017, huang_3d-vitac_2025}—often producing visually plausible yet globally inconsistent shapes.

In contrast, advances in 3D reconstruction and graphics demonstrate that dense, semantically meaningful representations can be recovered with minimal supervision. Neural radiance fields and related models, as well as 3D and 4D Gaussian Splatting, enable high-fidelity reconstruction and real-time rendering without object-specific retraining at inference \cite{li_neuralangelo_2023, kerbl_3d_2023, wu_4d_2024}. Meanwhile, structure-aware deformation models, including cage-based deformation and mass–spring-inspired formulations, provide compact control over dense geometry while naturally enforcing spatial smoothness and global coherence. These developments suggest that coupling local deformation cues with structured deformation propagation offers a principled path toward consistent and generalizable global shape reconstruction.

Motivated by these observations, we propose a zero-shot deformation sensing and reconstruction framework for soft robots based on structured geometric deformation modeling.
In our approach, a pre-trained flexible sensor patch can be directly attached to the surface of an unseen soft robot, without requiring robot-specific retraining or sensor layout optimization.
Local deformation signals captured by the sensor are interpreted as geometric boundary constraints and propagated to the entire robot body through a structure-aware deformation model that enforces spatial smoothness and continuity.
The resulting global deformation is represented using Gaussian-based primitives, enabling real-time, photorealistic RGB rendering without any camera input at inference time.

By decoupling local sensing, global deformation propagation, and visual rendering, the proposed framework achieves true zero-shot deployment while maintaining geometric consistency and real-time performance.

\medskip
\noindent\textbf{Contributions.}
The main contributions of this work are:
\begin{itemize}
    \item A zero-shot framework for soft robot deformation reconstruction that generalizes to unseen robot geometries without robot-specific pre-training.
    \item A structure-aware deformation propagation mechanism that expands sparse local sensor measurements into globally coherent 3D shape changes by enforcing spatial smoothness and continuity.
    \item A real-time Gaussian-based rendering pipeline that produces photorealistic RGB views of soft robot motion in camera-denied environments.
\end{itemize}

\section{Related Work}

\subsection{Sensor and Vision-Based Deformation Reconstruction}

Reconstructing the 3D geometry of deformable soft robots has been explored using both sensor-based and vision-based approaches. 
Early sensor systems employed sparse resistive or capacitive threads embedded in gloves \cite{delpreto_wearable_2022, sundaram_learning_2019, park_stretchable_2024, dong_modular_2025} or liquid-metal traces in soft actuators \cite{truby_soft_2019, truby_distributed_2020}, primarily estimating low-dimensional proxies such as joint angles rather than dense geometry.

More recent distributed e-skins based on fiber Bragg gratings, carbon nanotube composites, or dense strain arrays \cite{lun_real-time_2019, chen_structure_2024, mei_high-density_2024} improve local resolution, yet typically reconstruct deformation only on sensed regions through kinematic or regression mappings, limiting global consistency and visual fidelity.

EIT provides denser conductivity estimation \cite{chen_multi-touch_2025, dong_learning-enhanced_2025, hardman_multimodal_2025}, but remains layout-specific, calibration-intensive, and computationally demanding. Such systems often require object-specific configurations and struggle with zero-shot generalization, while real-time rendering remains challenging \cite{koch_surface_2017, zhao_measuring_2023}. Learning-enhanced skins can achieve millimeter-level accuracy \cite{hu_stretchable_2023}, though most depend on external ground truth during development.

Vision-based methods leverage multi-view stereo or SLAM to recover dynamic meshes or point clouds, while neural representations such as NeRF improve non-rigid modeling fidelity. However, these approaches require dense RGB input, precise calibration, and favorable lighting, making them brittle under occlusion or restricted viewpoints. Even recent particle–grid \cite{zhang_particle-grid_nodate} and neural field methods \cite{jiang_neurogauss4d-pci_nodate} remain difficult to deploy in high-resolution, real-time robotics scenarios without camera access.

\subsection{4D Gaussian Splatting}

4D Gaussian Splatting (4DGS) models dynamic scenes using temporally parameterized anisotropic 3D Gaussians. Compared to NeRF-based approaches, it enables high-fidelity reconstruction with real-time rendering and trackable primitives \cite{wu_4d_2024, kerbl_3d_2023, chen_survey_2025}.

Recent extensions improve geometric consistency or introduce structural constraints. TrimGS enhances surface quality via Gaussian pruning \cite{fan_trim_2024}, while GeoGaussian \cite{li_geogaussian_2024} and GausSurf \cite{wang_gaussurf_2024} enforce geometric priors. Sparse-control variants such as SC-GS \cite{huang_sc-gs_2024} and NeuroGauss4D-PCI \cite{jiang_neurogauss4d-pci_nodate} reduce motion complexity, and additional works incorporate deformation regularization \cite{xie_physgaussian_2024, abou-chakra_physically_nodate}.

Nevertheless, existing 4DGS pipelines depend on multi-view RGB supervision and accurate camera poses at inference, and stronger constraints often increase computational cost, limiting applicability in real-time, sensor-driven robotics.

\subsection{Cage-Based Deformation}

Predicting motion for dense meshes or Gaussian primitives is computationally expensive. Cage-based deformation introduces sparse control vertices whose motion drives high-resolution geometry, achieving global smoothness with low cost through harmonic or mean-value coordinates.

Recent work binds Gaussians to sparse cage nodes \cite{tong_cage-gs_2025, huang_gsdeformer_2025, huang_sc-gs_2024}, enabling efficient inference and editing by decoupling low-dimensional control from dense representation. This separation is particularly suitable for sensor-driven reconstruction: instead of regressing dense deformation fields, models infer sparse cage motion from sensor input and propagate it geometrically. Compared to array-based sensing methods that focus only on surface deformation or single deformation modes, cage-driven propagation supports globally consistent 3D reconstruction and naturally facilitates zero-shot transfer and real-time inference.

\begin{figure}[t]
    \centering
    \includegraphics[width=1.0\linewidth]{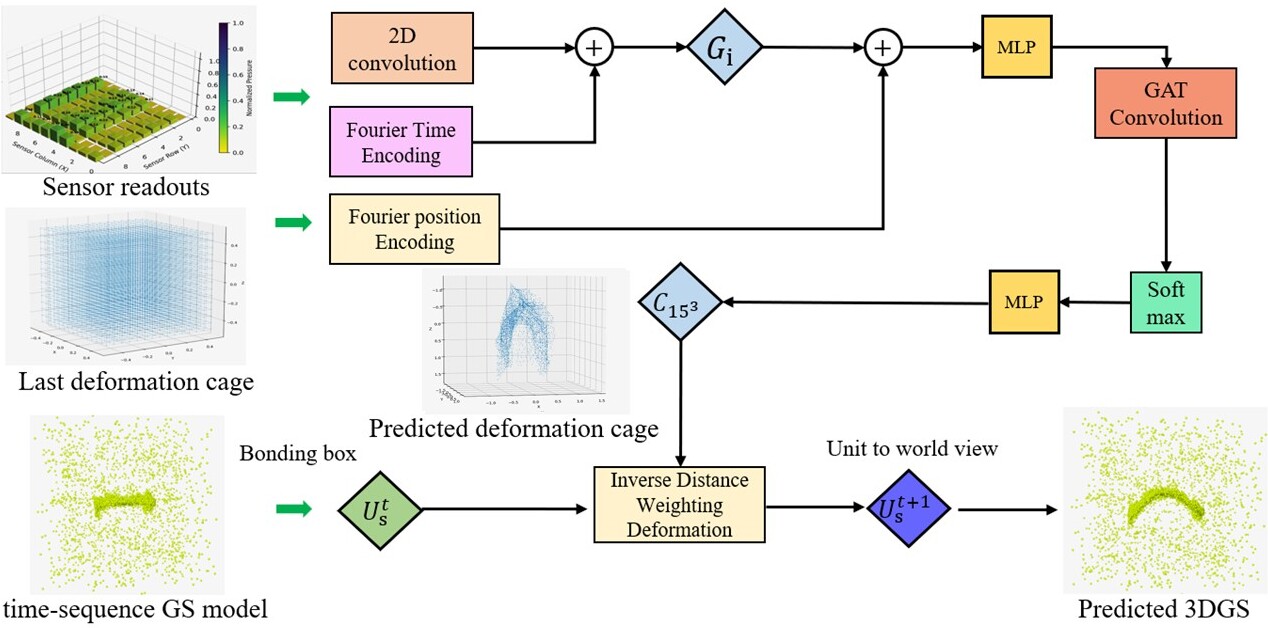} 
    \caption{The architecture of cage-based deformation networks. The cage is represented as a graph, where nodes correspond to cage vertices and edges encode local geometric adjacency. A graph attention (GAT) mechanism is employed to adaptively weight neighboring node contributions during deformation propagation, allowing the network to selectively emphasize structurally relevant neighbors under different deformation patterns. This attention-based message passing enables spatially coherent yet flexible deformation modeling, capturing non-uniform and anisotropic deformation behaviors across the cage.}
    \label{fig:2}
\end{figure}

\section{Method}\label{sec:method}

Our framework consists of four components:
(A) temporally consistent Gaussian supervision construction,
(B) flexible sensor hardware and signal acquisition,
(C) cage-based deformation prediction(Fig.~\ref{fig:2}), and
(D) real-time Gaussian rendering.
An overview is shown in Fig.~\ref{fig:1}.

\subsection{Temporally Consistent Gaussian Supervision Construction}

To obtain dense and temporally aligned supervision for learning the sensor-to-deformation mapping, we construct a dynamic Gaussian dataset in three stages.

\subsubsection{Multi-View Keyframe Capture}

For each deformation type (bending, twisting), we capture 20 key poses using a structured-light 3D scanner.
Each reconstructed mesh is rendered from 100 virtual viewpoints in Blender with fixed intrinsic and extrinsic parameters consistent with the 4DGS coordinate convention.
During capture, objects are rigidly fixed, the background is textureless, and the camera coordinate frame remains constant to ensure temporal alignment.

\subsubsection{Frame Interpolation}

To increase temporal density without excessive manual capture, we apply a pretrained interpolation network (RIFE~\cite{huang_real-time_2022}) to synthesize 8--16 intermediate frames between adjacent key poses for each view.
This results in approximately 100 views $\times$ 160 frames per motion sequence, providing smooth temporal coverage for subsequent 4DGS training.

\subsubsection{4D Gaussian Splatting Training}

The synthesized multi-view videos and camera poses are used to train a 4DGS model~\cite{wu_4d_2024}, producing temporally corresponded Gaussian primitives across frames.
Training follows the standard 4DGS configuration with RGB and SSIM losses, together with mild opacity and temporal smoothness regularization.
Depending on surface complexity, models converge with $5\times10^5$ to $2\times10^6$ Gaussians.
The resulting temporally aligned Gaussian centers serve as geometric supervision for cage-label extraction.

\subsection{Flexible Array Sensor Design and Data Collection}

\begin{figure}[t]
    \centering
    \includegraphics[width=1.0\linewidth]{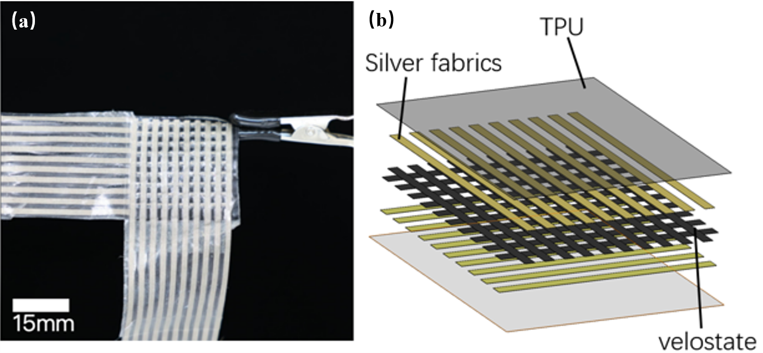}
       \caption{
        Tileable tactile sensor patch.
            (a) Photograph of the fabricated $3\times3$\,cm flexible tactile patch.
            (b) Exploded view illustrating the three-layer structure, consisting of
            (i) a $10\times10$ orthogonal grid of silver-plated nylon electrodes
            (ii) a Velostat piezoresistive layer,
            and (iii) TPU encapsulation.
            }
    \label{fig:sensor}
\end{figure}

We fabricate a $3\times3$\,cm tileable tactile patch composed of three layers following prior literature \cite{luo_tactile_2024, huang_3d-vitac_2025}:
(i) silver-plated nylon electrodes arranged in a $10\times10$ orthogonal grid,
(ii) a Velostat piezoresistive sheet,
and (iii) TPU encapsulation (Fig.~\ref{fig:sensor}).

Two CD74HC4067 multiplexers scan row and column electrodes within a voltage-divider circuit, and an ESP32 microcontroller digitizes the 100 channels (12-bit resolution, 200--500\,Hz sampling rate) into a $10\times10$ resistance map per frame.
A one-time two-point calibration maps raw voltages to resistances using per-channel baseline $R_0$.
During acquisition, a low-pass IIR filter (8--12\,Hz cutoff) and a $3\times3$ median filter are applied for temporal and spatial denoising.
For tiled configurations, patches share a synchronization line and distinct I\textsuperscript{2}C addresses, enabling timestamp-aligned concatenation into an $H\times W$ resistance image.
Each patch is aligned to the corresponding cage region via Procrustes fitting of four fiducial markers.

\subsection{Cage-Based Real-Time Deformation Prediction}

To bridge sparse sensor measurements and dense Gaussian primitives, we adopt a low-dimensional cage representation.

\subsubsection{Cage Representation and Gaussian Binding}

Each sensor region is enclosed by a regular-grid cage defined in a normalized canonical coordinate frame aligned with the local surface normal.
For each Gaussian center $p_j$, we precompute interpolation weights $w_{ij}$ with respect to its $k$ nearest cage nodes $c_i$.
We use inverse-distance weighting (IDW) with normalization:

\begin{align}
w_{ij} = \frac{\frac{1}{\|p_j - c_i\|+\varepsilon}}{\sum_{i' \in \mathcal{N}_k(j)} \frac{1}{\|p_j - c_{i'}\|+\varepsilon}}.
\end{align}

At runtime, cage displacements $\Delta c_i$ propagate to Gaussian centers as:

\begin{equation}
\Delta p_j = \sum_{i \in \mathcal{N}_k(j)} w_{ij} \Delta c_i .
\end{equation}

\subsubsection{Graph Attention Cage Deformer}

A lightweight CNN encodes the $H\times W$ sensor grid into a feature vector.
A temporal encoding (six-band sinusoidal embedding of normalized time) is concatenated to this feature.
The fused representation conditions a Graph Attention Network (GAT) operating on the cage graph (6-neighborhood connectivity).
The network contains two GAT layers (4 heads, 256 hidden units) followed by one GraphConv layer and an MLP projection to 3D node offsets.

\paragraph{Ground-Truth Cage Extraction}

Given Gaussian displacements $\Delta p_j^{\mathrm{gt}}$ from 4DGS and fixed weights $w_{ij}$, cage labels are computed by solving a per-frame least-squares problem:

\begin{equation}
\min_{\{\Delta c_i\}} \sum_j \left\|
\Delta p_j^{\mathrm{gt}} - \sum_i w_{ij} \Delta c_i
\right\|_2^2 .
\end{equation}

\paragraph{Training Objective}

The training loss consists of node-wise regression and temporal smoothness:

\begin{equation}
\mathcal{L} =
\frac{1}{N_c} \sum_i
\|\Delta c_i^{\mathrm{pred}} - \Delta c_i^{\mathrm{gt}}\|_2^2
+ \lambda
\|\Delta c_i^{\mathrm{pred}}(t) - \Delta c_i^{\mathrm{pred}}(t-1)\|_2^2 .
\end{equation}

At inference, exponential moving average (EMA) smoothing is optionally applied to improve temporal stability.

\subsection{Real-Time Gaussian Rendering}

Deformed Gaussians are rendered using a 3DGS-style rasterizer \cite{kerbl_3d_2023}.
Only Gaussian centers are updated per frame, while covariance, opacity, and color remain fixed from 4DGS training, enabling interactive rates.
Rendering resolutions range from $1280\times800$ to $1600\times1000$.
Near/far clipping, tile-based depth sorting, and frustum culling are enabled to reduce overdraw.
A lightweight UI allows adjustment of camera orbit, FOV, exposure, and EMA coefficient for live visualization.

\section{Experiments}

\subsection{Experimental Setup and Dataset}

Our experimental setup has two stages:
(1) controlled deformation modeling on a compliant substrate for supervised training,
and (2) zero-shot deployment on unseen soft robots.

\paragraph{Training platform on compliant substrate}

We first attach the $3\times3$\,cm tactile patch to the center of a $10\times10$\,cm highly compliant fabric sheet.
The fabric serves as a controlled deformable platform that allows repeatable bending and twisting motions while maintaining stable sensor contact.

For each deformation type (bending and twisting), we use a mechanical clamping tool to actuate the fabric.
By rotating the clamping position, we vary the bending or twisting central axis, generating multiple deformation configurations.
This setup allows systematic coverage of different curvature magnitudes and axis orientations, while ensuring that the sensor remains tightly coupled to the substrate.

\paragraph{High-fidelity 3D data acquisition}

For each motion type and each clamping orientation,
we capture 20 representative key poses spanning the full actuation range.
High-resolution textured meshes are reconstructed using an \textit{EinScan SE} structured-light scanner.
From each key pose, 100 virtual camera views are rendered in Blender with fixed intrinsic and extrinsic parameters.

To enrich temporal continuity, transitions between adjacent key poses are densified using video frame interpolation.
The resulting sequences contain approximately $1.6\times10^4$ images per motion type.
These multi-view sequences are processed using a 4DGS pipeline to obtain temporally consistent Gaussian representations.
These Gaussian sequences serve solely as supervision during training.

\paragraph{Sensor synchronization}

During data collection, resistance matrices from the tactile array are continuously recorded and synchronized with the corresponding 4DGS frames via timestamp alignment.
Raw resistance values are normalized per channel relative to their resting measurements.
Both cage control nodes and Gaussian centers are transformed into the canonical region coordinate frame, enabling region-wise training and later zero-shot transfer.

\subsection{Implementation and Training Details}

\paragraph{Cage-based deformation modeling}

Each sensor-covered region is modeled by a regular $15\times15\times15$ cage grid ($N_c=3375$).
The network predicts cage node displacements from sensor inputs, which are then propagated to Gaussian primitives via spatially weighted interpolation.
This formulation expands local sensor measurements to the full geometry while enforcing spatial smoothness and structural continuity, allowing unsensed regions to deform through structure-aware propagation rather than per-element regression.

\paragraph{Training protocol}

The model is implemented in PyTorch and PyTorch Geometric and trained solely on the compliant fabric substrate.
We train for 100 epochs using Adam (initial learning rate $1\times10^{-3}$ with cosine decay) and apply early stopping based on validation loss.
Gaussian noise is added to sensor inputs, and a temporal smoothness regularization term is imposed on cage displacements to enhance robustness.
All experiments are conducted on a single NVIDIA RTX 3070 Ti GPU.
\subsection{Zero-Shot Deployment on Unseen Soft Robots}

\begin{figure}[t]
    \centering
    \includegraphics[width=\columnwidth]{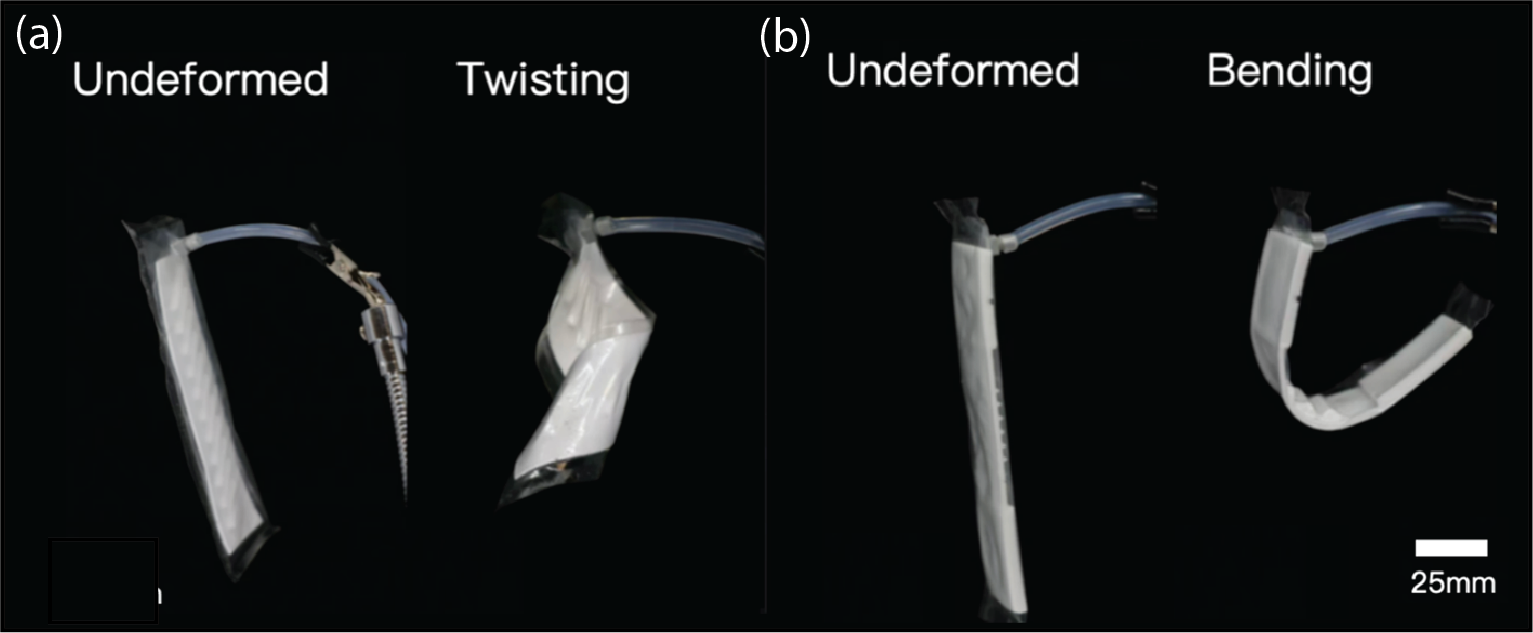}
    \caption{Two pneumatically actuated soft robots for zero-shot evaluation: (a) twisting and (b) bending.}
    \label{fig:example}
\end{figure}

To evaluate zero-shot generalization, we fabricate two soft robots that are not seen during training.
One robot undergoes pneumatic bending, while the other performs pneumatic twisting motions (Fig.~\ref{fig:example}). These deformation modes are selected as representative motion primitives commonly observed in soft robotic systems. For both robots, the undeformed geometry is initialized directly from the STL files generated during fabrication. No sensor-specific retraining, geometry-specific fine-tuning, or additional adaptation is performed prior to deployment.

\begin{figure*}[t]
    \centering
    \includegraphics[width=1.0\linewidth]{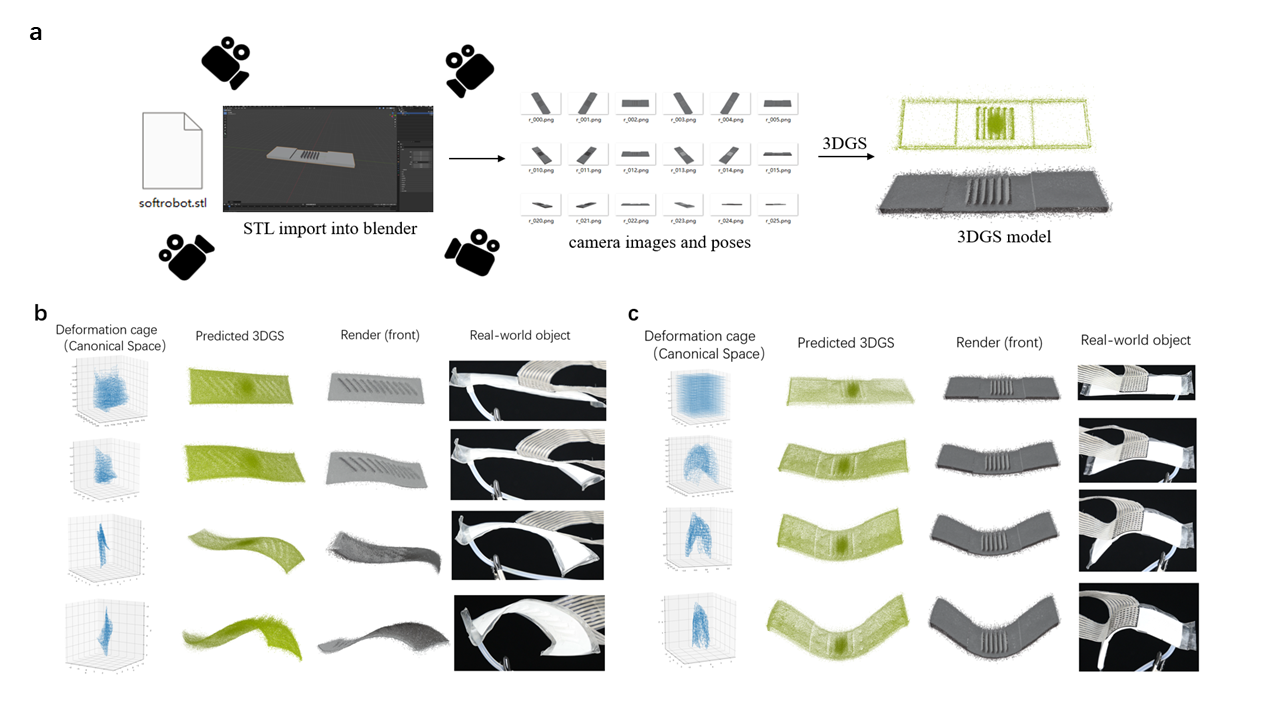}
    \caption{
    3DGS initialization and qualitative results.
    (a) Overview of the proposed geometry proxy initialization
    (b) Qualitative comparison for pneumatic twisting on an unseen soft robot.
    (c) Qualitative comparison for pneumatic bending on an unseen soft robot.
    Each comparison shows the predicted deformation cage, the reconstructed 3D Gaussian model (with viewpoints slightly rotated to better reveal surface indentations and protrusions), the rendered RGB view, and the corresponding real-world observation.
    }
\label{fig:fig5}
\end{figure*}

During inference, the same pre-trained tactile patch is attached to the central region of each robot. The recorded resistance matrices under pneumatic actuation are directly fed into the trained network to predict cage node displacements. These displacements deform the Gaussian representation in real time, while neighboring regions are updated through cage-based propagation to preserve global geometric coherence. Figure~\ref{fig:fig5} presents the initialization procedure and qualitative zero-shot results. 

Throughout the zero-shot evaluation, no external cameras or rendered supervision are used during inference. 
These results demonstrate that the proposed framework transfers from a fabric-based training platform to previously unseen soft robotic geometries without retraining, while maintaining coherent object-wide deformation reconstruction.

\section{Results}

We evaluate the proposed framework under a zero-shot setting, focusing on reconstruction accuracy, geometric consistency, robustness to complex surface geometry, and runtime performance on unseen soft robots.
Unless otherwise stated, all quantitative results are computed on held-out test sequences and averaged over time.

\subsection{Zero-Shot Reconstruction Accuracy}

\paragraph{Evaluation metrics}

Quantitative results are summarized in Table~\ref{tab:quant_results} using three complementary metrics: Intersection-over-Union (IoU), which measures global volumetric overlap between reconstructed and ground-truth 3D geometry; Chamfer Distance (CD), which evaluates local geometric accuracy via bidirectional nearest-neighbor distances between point sets; and Structural Similarity Index (SSIM), which assesses perceptual and structural consistency between rendered RGB images and ground truth.

\paragraph{Quantitative zero-shot results}

As reported in Table~\ref{tab:quant_results}, the proposed method achieves strong zero-shot reconstruction accuracy on both bending and twisting motions.
For the sensor-covered central region, reconstruction accuracy reaches \emph{2.18}\,mm Chamfer Distance and an IoU of \emph{0.76}.
When evaluated over the entire soft robot body, the method maintains robust performance with an average Chamfer Distance of \emph{3.48}\,mm and IoU of \emph{0.67}.

Among the two motion types, bending consistently yields higher accuracy than twisting.
The increased error in twisting is mainly observed in regions not directly covered by the sensor, where torsional motion introduces deformation patterns that are only indirectly constrained by the cage-based representation.
Despite this challenge, the framework produces globally coherent reconstructions without retraining, demonstrating effective geometric generalization under a zero-shot setting.

\subsection{Ablation Study on Deformation Representation}

We conduct an ablation study to evaluate key design choices in the deformation pipeline: (1) cage-based control versus direct dense regression, and (2) graph attention versus multilayer perceptron (MLP) for predicting cage displacements. All variants are trained and tested under the same zero-shot setting with identical sensor inputs and data.

\paragraph{Cage-based vs. Direct Regression}
Removing the cage representation and directly regressing per-Gaussian displacements leads to significantly higher geometric error and reduced stability, especially in regions not directly constrained by sensors. Without a low-dimensional control structure, local prediction errors accumulate along the elongated body, causing global drift and spatial inconsistency. These results highlight the importance of cage-based control in enforcing smoothness and global coherence under zero-shot deployment.

\paragraph{GAT vs. MLP for Cage Regression}
Replacing the graph attention network (GAT) with an MLP that independently predicts cage node displacements results in noticeable performance degradation, particularly in unsensed areas. By modeling adjacency and structural relationships within the cage graph, GAT enables coordinated motion among neighboring control nodes, leading to more consistent global reconstruction under complex deformations.

\begin{table}[t]
\centering
\caption{Ablation Study on Deformation Representation (Full Robot)}
\label{tab:ablation_combined}
\setlength{\textfloatsep}{8pt}
\setlength{\intextsep}{8pt}
\setlength{\tabcolsep}{6pt}
\renewcommand{\arraystretch}{1.15}

\begin{tabular}{|l|c|c|}
\hline
\multicolumn{3}{|c|}{\textbf{Geometric Accuracy}} \\
\hline
\textbf{Method} & \textbf{IoU} $\uparrow$ & \textbf{Chamfer (mm)} $\downarrow$ \\
\hline
Full Model (Cage + GAT + IDW) & \textbf{0.67} & \textbf{3.48} \\
w/o Cage (Direct Regression) & 0.41 & 4.62 \\
Cage + MLP (no GAT) & 0.62 & 3.97 \\
\hline
\end{tabular}

\end{table}

\subsection{Bending and Twisting Angle Accuracy}

\begin{figure}[t]
    \centering
    \includegraphics[width=1.0\linewidth]{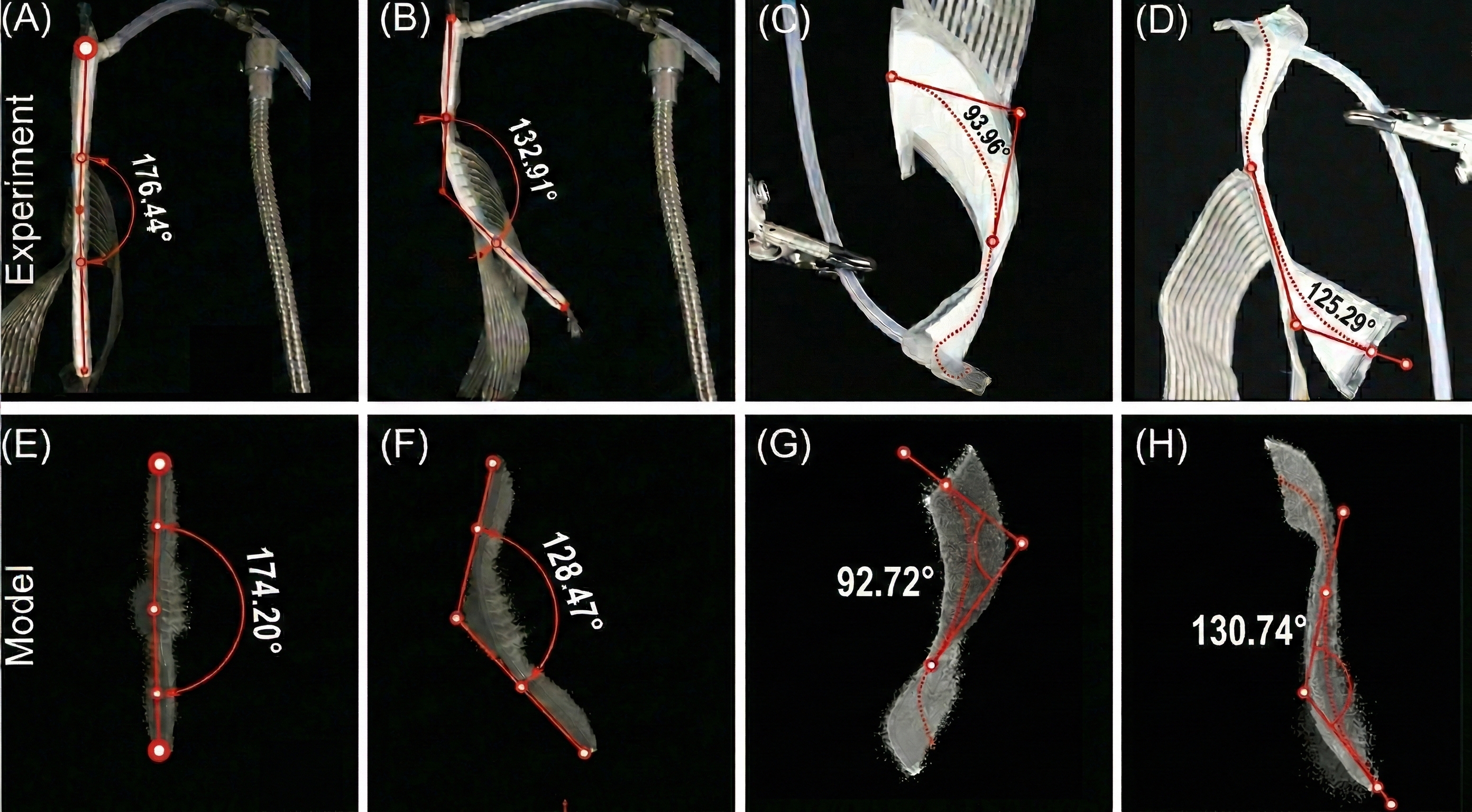}
    \caption{
    The top row (A–D) shows the real captured deformation scenes, while the bottom row (E–H) presents the corresponding results rendered in real time from the sensor. The deformation angles are manually measured and annotated using red lines and numerical labels. Specifically, A, B, E, and F correspond to bending cases, whereas C, D, G, and H correspond to twisting cases. Each top–bottom pair (A–E, B–F, C–G, D–H) represents the same deformation condition, enabling a direct comparison between the real captured angles and the sensor-based reconstruction results.
    }
    \label{fig:6}
\end{figure}

Beyond geometric metrics such as Chamfer distance and IoU, we additionally report motion-parameter-based evaluation to facilitate comparison with prior work. In particular, bending performance is quantified using the bending angle, which is the primary metric reported in existing literature. Across all bending sequences, the average angular error is \emph{4.7}$^\circ$. For twisting motions, where standardized quantitative metrics are less established, we adopt a centerline-based angular measure. Specifically, we compute the angle formed by the distal segments of the object’s centerline after twisting and compare it with real-world observations. The average twisting angular error is \emph{4.9}$^\circ$. Figure~\ref{fig:6} visualizes the angle comparisons between real-world measurements and rendered 3DGS results for both bending and twisting motions.

\subsection{Runtime Performance}

All experiments are conducted on a single NVIDIA RTX~3070\,Ti GPU.
The system operates at approximately \emph{30} frames per second (FPS) under the coarse Gaussian setting and \emph{5}\,FPS under the high-resolution setting.
The cage-based representation allows efficient control of dense geometry through a compact set of control nodes, enabling real-time deployment.
The visual and temporal differences between the coarse and high-resolution settings are demonstrated in the accompanying supporting video.

\subsection{Comparison with Prior Work}

Table~\ref{tab:comparison} summarizes the comparison with representative deformation reconstruction methods.

Vision-based approaches achieve high geometric accuracy under dense multi-view supervision, but fundamentally depend on calibrated cameras, line-of-sight visibility, and scene-specific optimization at inference. As a result, they are unsuitable for camera-denied, occluded, or space-constrained scenarios and generally lack zero-shot transfer to unseen robot geometries.

EIT-based and tactile/glove-based systems enable camera-free and real-time operation, yet rely on fixed sensing layouts, object-specific calibration, or dedicated training. Their reconstruction is often limited either by ill-posed inversion or by low-dimensional deformation descriptors, restricting scalability and generalization.

In contrast, the proposed method uniquely combines camera-free operation, real-time inference, zero-shot deployment, and dense 3D geometric reconstruction with photorealistic visualization. Although our Chamfer distance (3.48\,mm) and bending angle error (4.7$^\circ$) are not the lowest reported, they are achieved under a substantially more challenging zero-shot setting on unseen soft robots, without retraining or robot-specific calibration. 

Overall, our framework offers a balanced trade-off between reconstruction fidelity, global geometric consistency, and deployment flexibility, making it particularly suitable for practical, zero-shot deformation reconstruction in real-world soft robotic systems.

\begin{table}[t]
\caption{Quantitative Results of Zero-Shot Deformation Reconstruction}
\centering
\setlength{\tabcolsep}{4.5pt}
\renewcommand{\arraystretch}{1.1}
\begin{tabular}{|c|c|c|c|}
\hline
\textbf{Deformation} 
& \textbf{IoU} $\uparrow$ 
& \textbf{SSIM} $\uparrow$ 
& \textbf{Chamfer (mm)} $\downarrow$\\
\hline
Bending (center)
& $\mathbf{0.76 \pm 0.05}$ 
& -- 
& $\mathbf{2.18 \pm 0.32}$ 
\\
\hline
Bending (full)
& $0.69 \pm 0.18$ 
& $0.692 \pm 0.172$  
& $3.12 \pm 0.86$    
\\
\hline
Twisting (center)
& $\mathbf{0.72 \pm 0.07}$ 
& -- 
& $\mathbf{2.47 \pm 0.44}$ 
\\
\hline
Twisting (full)
& $0.65 \pm 0.21$  
& $0.603 \pm 0.186$ 
& $3.84 \pm 1.05$   
\\
\hline
\textbf{Average (full)}
& $\mathbf{0.67 \pm 0.20}$ 
& $\mathbf{0.65 \pm 0.18}$ 
& $\mathbf{3.48 \pm 0.96}$ 
\\
\hline
\end{tabular}
\label{tab:quant_results}
\end{table}

\begin{table*}[t]
\centering
\caption{Comparison with Existing Deformation Reconstruction Methods}
\label{tab:comparison}
\setlength{\textfloatsep}{8pt}
\setlength{\intextsep}{8pt}
\setlength{\tabcolsep}{4pt}
\renewcommand{\arraystretch}{1.1}
\begin{tabular}{|c|c|c|c|c|c|c|}
\hline
\textbf{Method} 
& \textbf{Occlusion Robust} 
& \textbf{Low-Latency} 
& \textbf{Zero-Shot} 
& \textbf{High-Fidelity Geometry} 
& \textbf{Angle Error ($^\circ$)} 
& \textbf{Chamfer (mm)}  \\
\hline
Vision-based \cite{fan_trim_2024}
& \ding{55} 
& \ding{55} 
& \ding{55} 
& \ding{51} 
& -- 
& $<$1.20 
\\
\hline
EIT-based \cite{chen_robot_2025,hu_stretchable_2023}
& \ding{51} 
& \ding{51} 
& \ding{55} 
& \ding{55} 
& 1.90 
& 2.32
\\
\hline
Tactile / Glove-based \cite{park_stretchable_2024}
& \ding{51} 
& \ding{51} 
& \ding{55} 
& \ding{55} 
& 4.16 
& 4.02
\\
\hline
\textbf{Ours} 
& \ding{51} 
& \ding{51} 
& \ding{51} 
& \ding{51} 
& \textbf{4.7} 
& \textbf{3.48}  \\
\hline
\end{tabular}
\end{table*}

\section{Discussion}

This work presents a camera-free, zero-shot deformation reconstruction framework that couples flexible tactile sensing with a structured cage-driven Gaussian representation. By decoupling sparse control from dense geometry, the system enables real-time inference while preserving global geometric coherence and high-fidelity rendering, providing a practical pathway toward scalable digital twin deployment in occluded or low-visibility environments.

A key observation is the trade-off between real-time performance and physically accurate deformation modeling. To maintain low latency, the framework adopts structure-aware geometric propagation rather than explicit material simulation. While this design ensures stable and efficient inference, reconstruction accuracy may degrade under extreme deformations or complex contact conditions, particularly when nonlinear material behavior dominates the response. This reflects a broader balance between computational efficiency and physics fidelity in sensor-driven reconstruction systems.

We also observe that the IDW scheme used for deformation propagation reduces overlap accuracy in regions with abrupt curvature transitions, such as tightly curled endpoints. These high-curvature regions involve localized stress concentration that cannot be fully captured through purely geometric interpolation, suggesting the need for material-dependent physical priors in highly nonlinear regimes.

Reconstruction quality is additionally sensitive to sensor–surface conformity, with imperfect attachment or localized interactions (e.g., sharp pressing) introducing measurement inconsistencies. Each new object further requires an offline geometric initialization step to construct its canonical Gaussian representation, resulting in a modest setup overhead.

Looking ahead, localized physics-assisted reconstruction—such as integrating Material Point Method (MPM) simulation in high-curvature regions—could enhance fidelity while preserving global efficiency. The cage-based formulation also supports modular scaling: multiple sensor patches can be stitched or hierarchically organized through coordinated cage regions, enabling distributed sensing without fundamentally increasing complexity. Incorporating adaptive material parameter estimation may further improve robustness across diverse soft substrates.

Overall, coupling tactile sensing with structured geometric deformation propagation provides a scalable foundation for real-time, zero-shot soft robot reconstruction, with clear opportunities for physics integration and modular expansion.
Beyond soft robotics, the proposed approach can extend to richer contact scenarios and everyday objects. As a preliminary demonstration, Fig.~7 shows a proof-of-concept experiment in which the sensor is attached to a wallet to reconstruct pressure-induced deformation in real time, highlighting the potential of this framework for embodied manipulation and interactive applications.

\begin{figure}[t]
    \centering
    \includegraphics[width=1.0\linewidth]{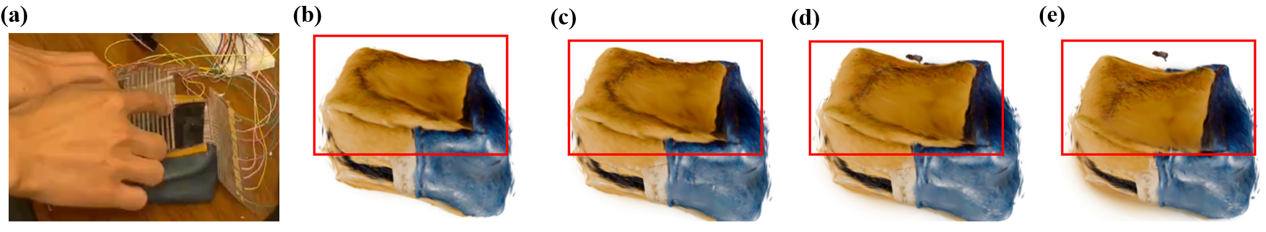}
    \caption{
    Proof-of-concept tactile deformation reconstruction on an everyday object.
    (a) A human hand presses a wallet instrumented with a surface-mounted flexible tactile sensor.
    (b--e) Time-sequential renderings of the reconstructed 3DGS model, showing progressive deformation under increasing pressure.
    The results demonstrate real-time pressure-to-geometry reconstruction beyond soft robotic platforms.
    }
    \label{fig:wallet}
\end{figure}
\section{Conclusion}

We presented a zero-shot, camera-free deformation reconstruction framework that integrates flexible tactile sensing with a cage-based geometric deformation model and Gaussian rendering. Although the method does not achieve the lowest geometric error compared to robot-specific approaches, it demonstrates strong zero-shot generalization by explicitly coupling local sensor measurements with structured deformation propagation, enabling real-time, globally consistent reconstruction without vision at inference.

The results suggest that combining tactile sensing with expressive geometric representations and deformation priors provides a scalable foundation for transferable shape reconstruction. Future work will investigate physics-assisted deformation modeling and adaptive material-aware representations to further improve fidelity while preserving real-time performance. Owing to its modular cage-based formulation, the framework also supports scalable multi-sensor integration and broader deployment.


\balance
\bibliographystyle{IEEEtran}
\bibliography{references_short}

\end{document}